\documentclass[runningheads]{llncs}
\usepackage{graphicx}
\usepackage{color}
\usepackage{array}
\usepackage[font={footnotesize}]{caption}
\usepackage{mathtools}
\usepackage{multirow}
\usepackage{subfig}
\usepackage{float}

\graphicspath{ {./figures/}}

\newcolumntype{C}[1]{>{\centering\let\newline\\\arraybackslash\hspace{0pt}}m{#1}}

\begin{document}
\title{Brain-inspired algorithms for processing of visual data}
%
%
\author{Nicola Strisciuglio}
\authorrunning{N. Strisciuglio}
%
\institute{Faculty of Electrical Engineering, Mathematics and Computer Science\\ University of Twente, The Netherlands\\
\email{n.strisciuglio@utwente.nl}}
\maketitle              
\begin{abstract}
The study of the visual system of the brain has attracted the attention and interest of many neuro-scientists, that derived computational models of some types of neuron that compose it. These findings inspired researchers in image processing and computer vision to deploy such models to solve problems of visual data processing.

In this paper, we review approaches for image processing and computer vision, the design of which is based on neuro-scientific findings about the functions of some neurons in the visual cortex. Furthermore, we analyze the connection between the hierarchical organization of the visual system of the brain and the structure of Convolutional Networks (ConvNets). We pay particular attention to the mechanisms of inhibition of the responses of some neurons, which provide the visual system with improved stability to changing input stimuli, and discuss their implementation in image processing operators and in ConvNets.

\end{abstract}
\section{Introduction}
The development of the visual system of humans takes a number of phases, which include tuning the synaptic connections between neurons in the different areas devoted to the processing of different visual stimuli. In newborns, for instance, many connections between the Lateral Geniculate Nucleus (LGN), which is the first part of the brain devoted to visual processing, and the area V1 of the visual cortex are not formed yet. Similarly, the connections between neurons in the area V1 and subsequent areas start developing after the first month of life.

The tuning process of the receptive fields of the neurons of the visual system and the development of their inter-connected network can be compared to the training process of Artificial Neural Networks (ANNs). Since the beginning of their development, indeed, the design of ANNs has been largely inspired by the way the brain works, i.e. processing information via a network of neurons organized in a hierarchical fashion. Despite the resemblance of the Rosenblatt's perceptron with the physiological structure of a neuron, there is no actual relation between the processing of ANNs and the neural processes in the brain.

Many researchers in computer vision and image processing  found inspirations from neuro-physiological studies of the visual system of the brain to design novel computational models that could process visual data. 
In 1959, Hubel and Wiesel carried out experiments on the visual cortex of cats and demonstrated the existence of the \emph{simple cells}, which are neurons with an elongated receptive field. Their primary function is to detect edges and lines. Originally, the simple cells were modeled using Gabor functions~\cite{daugman1985uncertainty,jones1987evaluation} and used in image processing and computer vision applications, especially for texture description and analysis~\cite{grigorescuTexture}. 
Subsequently, Hubel and Wiesel precised that simple cells receive inputs from certain co-linear configurations of the circular receptive field of neurons in the LGN~\cite{hubel1962receptive}. Computational models based on Gabor functions were not able to  describe all the properties of simple cells and ignored the contribution of LGN neurons for the processing of visual stimulti. In~\cite{azzopardi2012corf}, a computational model based on the combination of the responses of Difference-of-Gaussians functions, which modeled the LGN receptive fields, was proposed. It achieved better contour detection performance than models based on Gabor functions and showed more properties of the simple cells in area V1 of the visual system of the brain, such as contrast invariant orientation tuning and cross orientation suppression.

Artificial neural networks (ANNs) and, in particular, convolutional neural networks (ConvNets) received much attention and showed some similarities with the visual system of the brain especially regarding its hierarchical organization. Although the training of neural network is formulated as an optimization problem and does not relate with biological processes, in~\cite{alexnet} it was shown that the convolutional kernels learned in the first layer of AlexNet resembled the Gabor functions that were used to model the receptive field of neurons in the area V1 of the visual system. Similarly, unsupervised approaches for image analysis like Independet Component Analysis also learned features for image processing that resemble the Gabor-like receptive fields of neurons in area V1~\cite{ICA}.

Neuro-scientific and neuro-physiological studies of the mechanisms and systems that our brains uses to process external inputs have influenced also the developement of other branches of pattern recognition and artificial intelligence, such as sound signal processing. Patterson \emph{et al.}, in 1986, modeled the response of the cochlea membrane in the inner auditory system as a bank of Gammatone filters~\cite{patterson1986auditory}. They called Gammatonegram the result of the processing of an input signal by a Gammatone filter bank. Similarly to the spectogram, the Gammatonegram is a time-frequency representation of the sound in which the energy distribution over time and specific bandwidths is described. Parts of higher energy intensity correspond to regions of the cochlea membrane that vibrates more according to the energy of the mechanical sound pressure waves that hit the outer part of the auditory system. This model was exploited in~\cite{StrisciuglioCOPE,CopePreliminary2015,COPE2019} as input to a trainable feature extractor, the design of which was inspired by the activation of the inner hair cells, placed behind the cochlea, which convert the vibration into electrical stimuli on the auditory nerve.

This paper focuses on the relation between neuro-scientific studies and progress in Computer Vision and Image Processing, providing an overview of methods and aspects that concern detection and processing of low-level features in images until more complex computations in convolutional networks.

\section{Brain-inspired processing of visual data}
One of the pioneering architectures for image processing and computer vision inspired by knowledge of the brain processes of vision was the neocognitron network~\cite{Fukushima1980}. It modeled the hierarchical arrangement of the visual system of the brain by layers of S- and C-cell components, which are computational models of the simple and complex cells discovered by Hubel and Wiesel~\cite{hubel1962receptive}. The weights of the neocognitron network were learned via an unsupervised training process, based on self-organizing maps. This training resulted in a hierarchy of S- and C-cell units that resembled the organization of the human visual system. 


In the following of the section, some of these approaches are discussed, and part of the focus is given to the phenomena of inhibition that contribute to increase the selectivity of neurons to specific visual stimuli and how they are embedded in operators for processing of visual data.

\subsection{Edge and line detection}

Simple cells in area V1 of the visual cortex receive inputs from LGN cells in the thalamus of the brain and have the function of detecting elongated structures that contain high contrast information. The receptive fields of LGN cells are modeled by on- and off-center Difference-of-Gaussians (DoG) functions, while those of simple cells are modeled as co-linear arrangement of DoG functions. 
Originally, simple cells were modeled with Gabor functions, bypassing the contribution of the LGN cells. Computational models based on Gabor filters were used for contour and line detection and included in hierarchical architectures for object detection~\cite{serre2005object} and face recognition~\cite{pinto2011scaling} tasks. 

Although Gabor filters were used, initially, to model the simple cell receptive fields~\cite{jones1987evaluation}, they did not reproduce certain properties, such as contrast invariant orientation tuning and cross orientation suppression. 
These properties were achieved by a non-linear model, named CORF (Combination of Receptive Fields) for contour detection~\cite{azzopardi2012corf}. It is based on the combination of co-linearly aligned DoG functions, modeling the way simple cells combine the response of LGN cells. A mechanism for tolerance to curvature of lines and contours, based on a non-linear blurring, was proposed in the CORF model to improve the results when deployed in image processing pipelines.

\begin{figure}[!t]
\centering
	\setlength{\unitlength}{30mm}
	\subfloat[]{\label{fig:retina1}
		\includegraphics[height=\unitlength]{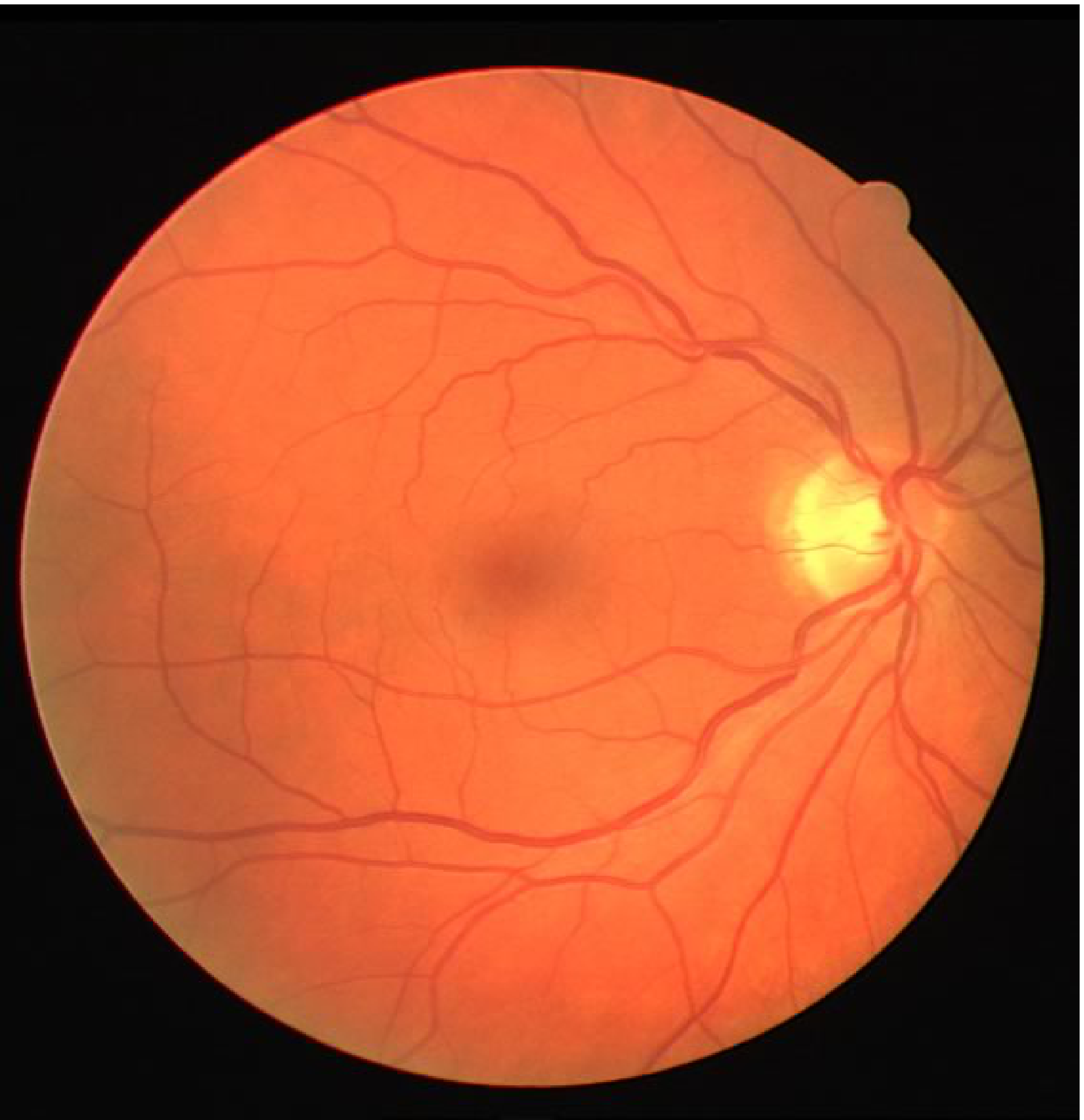}
	}
	\subfloat[]{\label{fig:retina3}
		\includegraphics[height=\unitlength]{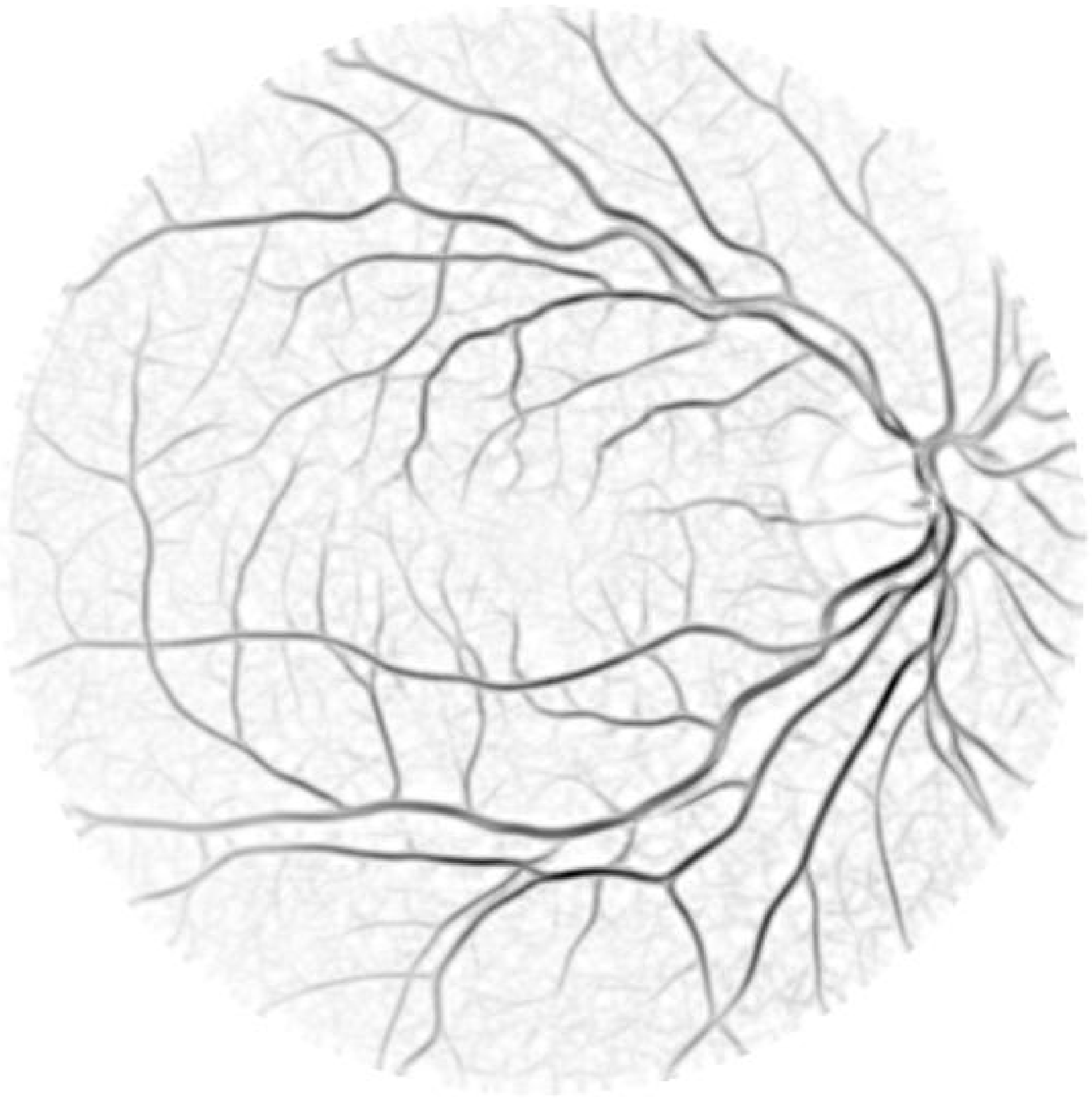}
	}	
	\subfloat[]{\label{fig:retina4}
		\includegraphics[height=\unitlength]{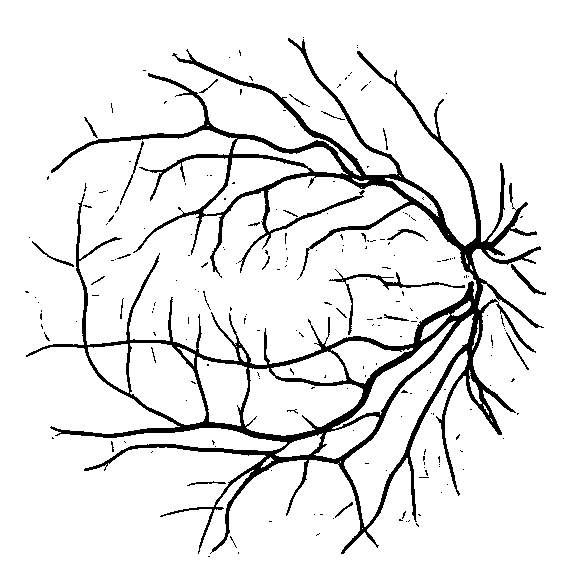}
	}
\caption{(a) Example retinal image, the (b) response of the B-COSFIRE filter and (c) the corresponding binary map.}
\label{fig:retina}
\end{figure}

An implementation of CORF, named (B-)COSFIRE (Combination of Shifted Filter Responses), where B- stands for bar-selective, was demonstrated to be successful for the detection of thick lines in images and applied to blood vessel delineation in retinal images (see Fig.~\ref{fig:retina})~\cite{StrisciuglioVIP15,AzzopardiStrisciuglio2015}, road and river segmentation in aerial images~\cite{strisciuglio2017delineation}, crack detection in pavement images~\cite{strisciuglio2017detection}. An example of the response map computed by a B-COSFIRE filter and its thresholded binary map are shown in Fig.~\ref{fig:retina}b and Fig.~\ref{fig:retina}c, respectively. A curved receptive field was configured in~\cite{Sivakumar2020}, to detect high curvature points of the retinal vessel tree. In~\cite{Strisciuglio2016,Strisciuglio15}, the authors demonstrated that a bank of B-COSFIRE filters, configured to delineate lines of different thickness, can be used as feature extractors and combined with a classifier to perform complex decisions.

\subsection{Object(-part) detection}
The response of neurons in area V1 are forwarded for further processing to neurons in areas V2 an V4 of the visual cortex, which are tuned to respond to sets of curved segments or vertices of some preferred orientation and badnwidth~\cite{Pasupathy12}. These properties can be interpreted as functions for detection of parts of objects. 

\begin{figure}[!t]
\centering
	\setlength{\unitlength}{30mm}
	\includegraphics[height=\unitlength]{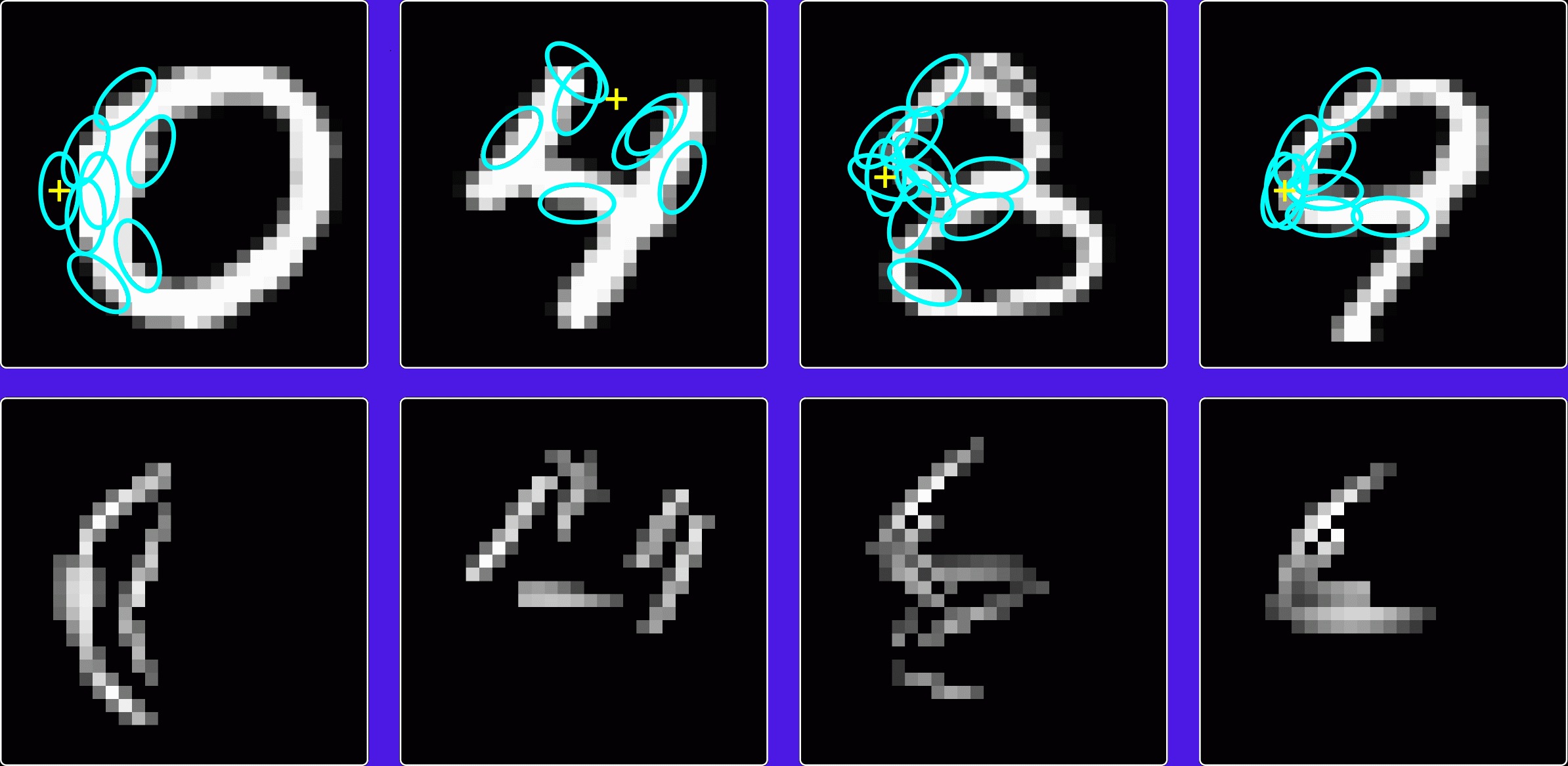}
\caption{The configured COSFIRE filters are represented by the set of light blue ellipses in the top row, whose orientation indicates the preferred orientation of the Gabor filter. In the bottom row, the part of the object that the corresponding COSFIRE filter is able to detect (figure from~\cite{COSFIRE}).}
\label{fig:v4}
\end{figure}

Based on the principle of combining the responses of line and edge detectors at different orientations and with a certain spatial arrangement, an implementation of the COSFIRE model that takes as input a bank of Gabor filters of different orientation was released~\cite{COSFIRE}. In this case, the receptive fields of neurons in area V1 that give input to those in area V4 were modeled by means of Gabor functions. However, a hierarchical structure of COSFIRE models can be realized for more complex tasks like object recognition or scene understanding~\cite{AzzopardiShape}.
 The COSFIRE model of neurons in area V4 can be trained to detect parts of object and used in applications of object recognition. In Fig.~\ref{fig:v4}, we show some examples of the parts of objects on which V4-COSFIRE models are trained. The light-blue ellipses indicate the location and the orientation at which the V1-like neuron responses are considered and their combination models a part of the object of interest. The configured models can be used to recognize parts of objects in other images or together in a filter-bank to extract feature vectors to be used in combination with a classifier.

\subsection{Inhibition for image processing}

One important aspect of the visual processes that happens in the visual system is the mechanism of inhibition. The receptive field of a simple cell, known as `classical receptive field'~\cite{hubel1959receptive}, is composed of an excitatory and an inhibitory region. Many simple cells are know to receive push-pull (or antiphase) inhibition~\cite{hubel1965receptive}. This form of inhibition happens when visual stimuli of given orientation and opposite polarity evoke responses of opposite sign~\cite{palmer1981receptive,borg1998visual,ferster1988spatially}. Furthermore, it is known to be the most diffuse form of inhibition in the visual cortex~\cite{anderson2000orientation}. In practice, for a stimulus of given polarity the response of the inhibitory receptive field suppresses the response of the excitatory receptive field.

\begin{figure}[!t]
\centering
	\setlength{\unitlength}{25mm}
	\subfloat[]{\label{fig:inhib1}
		\includegraphics[height=\unitlength]{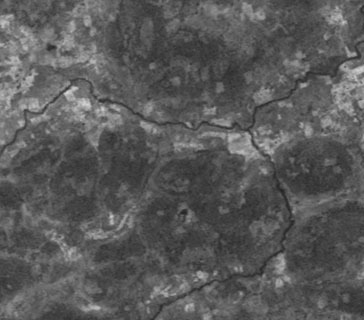}
	}
	\subfloat[]{\label{fig:inhib2}
		\includegraphics[height=\unitlength]{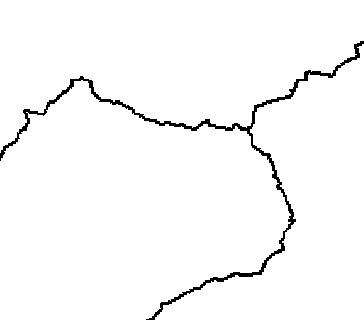}
	}	
	\subfloat[]{\label{fig:inhib3}
		\includegraphics[height=\unitlength]{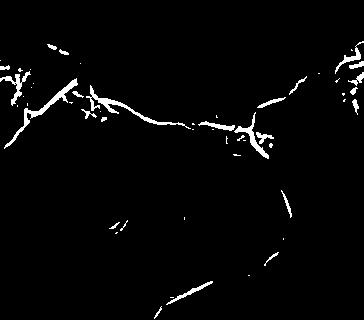}
	}	
	\subfloat[]{\label{fig:inhib4}
		\includegraphics[height=\unitlength]{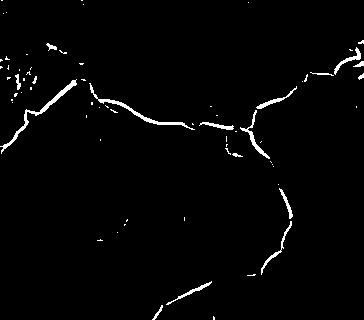}
	}
\caption{(a) Aerial image of a river and (b) the ground truth of the river area. The (c) binary response map obtained by the B-COSFIRE filter is more noisy and contains less of the detected river patterns than the (d) binary response map of RUSTICO.}
\label{fig:inhibexample}
\end{figure}

This phenomenon was implemented in the CORF operator and it was demonstrated to be beneficial for improving contour detection in presence of texture~\cite{azzopardi2014push}. More recently, the effect of the push-pull inhibition was shown to increase the robustness of line detection to various types of noise and textured background: a novel RUSTICO (Robust Inhibition-augmented curvilinear operator) operator was proposed in~\cite{StrisciuglioTIP2019,StrisciuglioECCV18}.  It was shown to be very effective for line detection in presence of noise and texture. RUSTICO is designed as an extension of the B-COSFIRE filter for line detection, by including an inhibitory component. In Fig.~\ref{fig:inhib1} and Fig.~\ref{fig:inhib2}, an aerial image of a river and the corresponding ground-truth are shown. The binary response map produced by RUSTICO (Fig.~\ref{fig:inhib4}) shows a more complete reconstruction of the line pattern of interest, i.e. the river, than that in the binary map produced by B-COSFIRE (Fig.~\ref{fig:inhib3}).

Another phenomenon of inhibition found in the visual cortex is the surround suppression. It consists of neurons, whose response is suppressed by that of neighbor neurons in the surrounding of their receptive field~\cite{bishop1973receptive,wiesel1966spatial}. The cells that exhibit this type of inhibition have a non-classical receptive field (NCRF). Practically, this means that the response to a certain stimulus can be influenced by the presence of similar stimuli in the surrounding of the receptive field. This mechanism of surround suppression was included in image processing operators to extend the Canny edge detector~\cite{grigorescu2003contour}, a Gabor filter based contour detector~\cite{grigorescu2004contour} and in an operator with a butterfly-shaped receptive field~\cite{zeng2011contour}.

More recently, the push-pull inhibition and surround suppression were combined in a single operator for contour detection, which outperformed its counterpart operators with single or none inhibition mechanism~\cite{Melotti2020}.

\section{Convolutional networks for visual data processing}
Convolutional Neural Networks (ConvNets) became the \emph{de facto} standard for image processing and computer vision, because of their effectiveness in dealing with various visual recognition tasks. Successful applications of ConvNets are image and object recognition~\cite{resnet}, semantic segmentation~\cite{segnet}, place recognition~\cite{netvlad,leyva2019}, image generation and image-to-image translation~\cite{pix2pix2017}, among others.

ConvNets are based on convolution operations and exploit the characteristic of locality of the patterns of interest. This means that the value at a certain pixel location of a response map is detemined by the linear combination of the values of a small neighborhood of the corresponding pixel in the input image. 
From this perspective, ConvNets can be considered as a regularized version of multi-layer perceptron (MLP) networks. The fully-connectedness means that each neuron at a certain layer receives input from all the neurons in the previous layer. In a ConvNet, instead, each neuron (i.e. a convolution kernel) has a very limited number of inputs, and it slides over the input signal to compute its response. Although a single convolution catches local proprieties of the input signal in small-size neighboroods, the hierarchical organization of ConvNets allows to assemble more and more complex patterns in subsequent steps.

\begin{figure}[!t]
\centering
	\setlength{\unitlength}{110mm}
	\includegraphics[width=\unitlength]{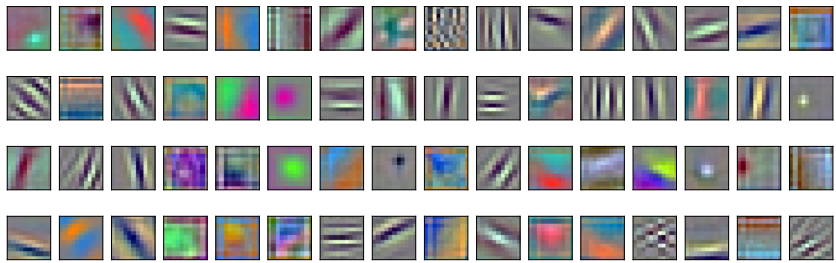}
\caption{Visualization of the convolutional kernel learned in the first layer of AlexNet.}
\label{fig:alexnet}
\end{figure}

The hierarchical organization of ConvNets, which arranges a stack of convolutional layers, non-linear activation functions and sub-sampling operations  resembles the hierarchy of the visual system of the brain. Speculations of this type were reinforced by the results obtained by the AlexNet network~\cite{alexnet}. On top of the improvement of the classification accuracy by a large margin with respect to previous approaches, it was shown that the filters learned in the first layer of AlexNet resembled Gabor-like receptive fields (see Figure~\ref{fig:alexnet}), which are accepted computational models of neurons in the area V1 of the visual system of the brain~\cite{Marcelja80}. 
Hence, in the first layer of AlexNet edge and elongated structures of different bandwidth are detected. The interpretations consist in that in subsequent layers, the detected edge and line patterns are combined into corner-like structures, similarly to the area V2 and V4 of the visual cortex, and into parts of objects (anterior and posterior TEO). 

The convolutions used in ConvNet architectures are linear operators and are not able to fully model some non-linear properties of the neurons in the visual cortex, e.g. response saturation or cross-orientation suppression. 
In~\cite{volterracnn}, quadradic convolutions, in the form of Volterra kernels, were investigated and deployed as substitute of the convolution operations in existing architectures. This type of convolutions is more suited for a better approximation of the profile of the receptive fields of some neurons in the visual system. The approach was extended in~\cite{Jiang2019}, in which quadratic convolutional kernels contributed to reduce the depth, i.e. the total number of convolutional layers, of existing architectures while keeping the detection and classification performance of the corresponding deeper original networks. 

On the one hand, the use of quadratic convolutions is justified by 
the closer connection with the function of the receptive field of the complex cells in the visual system, and contributed to a relatively small increase of performance. On the other hand, they require a much larger number of parameters to be learned, slowing down the training and increasing the complexity of the functions to be learned. In~\cite{volterracnn}, indeed, due to computational limits, only the first layer of convolutions was replaced by Volterra kernels.

Another type of non-linear unit was proposed in~\cite{LopezAccess2019}, which incorporate the framework of the COSFIRE model of the neurons in the area V4 of the visual system into a new type of layer for ConvNets. 
The response of this layer is computed by combining the response maps of local simpler features according to a spatial structure that is determined in an automatic configuration step. During the training of the network, the CNN-COSFIRE layer can be configured to detect a certain arrangement of local features, so allowing for a larger receptive field that can catch non-local characteristics of the patterns of interest, such as parts of or entire objects. It was successfully demonstrated in applications of object detection and place recognition where few training samples ara available.

\subsection{Inhibition in convolutional networks}
ConvNets learn representations, disentangling complex features of the training data. Inhibition is believed to be a mechanism for regularization and stability of the processes that happens in the visual system~\cite{Lauritzen10201}, and forms of inhibition are learned in ConvNets as well~\cite{Tjostheim2019}.

AlexNet deployed a layer called Local Response Normalizer (LRN), which implemented a surround suppression mechanism called lateral inhibition. This type of inhibition creates a form of competition among neurons in a local neighboround. The LRN builds on the idea of enhancing peak responses and penalizing flat ones on the feature map, making relevant features stand out more clearly. Thus, in the implementation, high local responses of one convolutional kernel inhibit weaker responses of other convolutional kernels in the same local neighbourhood. This serves as a form of regularization of the network and improves recognition performance. 

In~\cite{strisciuglio2020pushpull}, a new type of layer that implements the push-pull inhibition mechanism was proposed, which can be used as a substitute of the convolutional layer. The push-pull layer can be trained with back-propagation of the gradient of the error and is interchangeable with any convolutional layer in the network. However, as it is inspired by neuroscientific evidence of inhibition mechanisms that occur in the early stages of the visual cortex, it was deployed as a substitute of the first convolutional layer only~\cite{strisciuglio2020pushpull}.
Using the push-pull layer in ConvNet architectures achieves better performance on image classification tasks when dealing with images that have been corrupted with noise or other types of artefacts (e.g. jpeg compression, blur, contrast changes and so on). Furthermore, when deploying the push-pull layer in ConvNets instead of a convolutional layer, the number of parameters to learn does not increase.

\section{Conclusions}
\label{sec:conclusions}
The research fields of image processing and computer vision were influenced by discoveries and progress in the understanding of the functions of neurons in the visual system. Computational models of different types of neurons formalized by neuro-physiological studies of their responses to visual stimuli have been deployed for image processing, especially related to low-level tasks such as line and contour detection.

In this paper, we reviewed the developments of edge and contour detection algorithms influenced by progress made in the understanding of the visual processes that occur in the visual cortex. We paid large attention to the importance that inhibitory mechanisms, namely push-pull inhibition and surround suppression, have on the robustness of the processing of visual stimuli in noisy and textured scenes. 
Furthermore, we covered the connections that neuro-physiological findings hae with the development of Convolutional Networks and how inhibitory phenomena were explicitly implemented in the architecture of these networks with the aim of improving their stability to varying input stimuli.


\section*{Acknowledgments}
I would like to thank Maria Rosaria Strisciuglio for the interesting  discussions about the phases of learning and development of the visual system of the brain.

\bibliographystyle{splncs04}
\bibliography{citations}

\end{document}